\documentclass[conference]{IEEEtran}
\IEEEoverridecommandlockouts

\usepackage{cite}
\usepackage{amsmath,amssymb,amsfonts}
\usepackage{algorithmic}
\usepackage{graphicx}
\usepackage{textcomp}
\usepackage{multirow}
\usepackage{verbatim}
\usepackage{color}
\usepackage{xcolor}
\definecolor{hiddendraw}{rgb}{0.7,0.7,0.7}
\usepackage{tikz}
\usepackage[edges]{forest}
\def\BibTeX{{\rm B\kern-.05em{\sc i\kern-.025em b}\kern-.08em
    T\kern-.1667em\lower.7ex\hbox{E}\kern-.125emX}}
\begin{document}

\title{Med-PU: Point Cloud Upsampling for High-Fidelity 3D Medical Shape Reconstruction\\
}

\author{\IEEEauthorblockN{Tongxu Zhang\textsuperscript{1} and Bei Wang\textsuperscript{1}}
\IEEEauthorblockA{\textit{\textsuperscript{1} East China University of Science and Technology, Xuhui, Shanghai, China} \\
juk@mail.ecust.edu.cn, beiwang@ecust.edu.cn}
}

\maketitle

\begin{abstract}
High-fidelity 3D anatomical reconstruction is a prerequisite for downstream clinical tasks such as preoperative planning, radiotherapy target delineation, and orthopedic implant design. We present \textit{Med-PU}, a knowledge-driven framework that integrates volumetric medical image segmentation with point cloud upsampling for accurate pelvic shape reconstruction. Unlike landmark- or PCA-based statistical shape models, Med-PU learns an implicit anatomical prior directly from large-scale 3D shape data, enabling dense completion and refinement from sparse segmentation-derived point sets. The pipeline couples SAM-Med3D-based voxel segmentation, point extraction, deep upsampling, and surface reconstruction, yielding smooth and topologically consistent meshes. We evaluate Med-PU on pelvic CT datasets (MedShapePelvic for training and Pelvic1k for validation), benchmarking against state-of-the-art upsampling methods using comprehensive geometry and surface metrics. Med-PU consistently improves surface quality and anatomical fidelity while reducing artifacts, demonstrating robustness across input densities. Although validated on the pelvis, the approach is anatomy-agnostic and applicable to other skeletal regions and organs. These results suggest Med-PU as a practical, generalizable tool to bridge segmentation outputs and clinically usable 3D models.

\end{abstract}


\section{Introduction}
High-fidelity 3D anatomical reconstruction is a prerequisite for several downstream clinical tasks, including preoperative planning, radiotherapy target delineation, and orthopedic implant design. In routine pipelines, clinicians often begin with voxel-wise segmentation and then derive surface meshes; however, directly converting voxel outputs into smooth, watertight, and anatomically faithful meshes remains challenging due to partial observations, image noise, and inter-subject variability.

Traditional \textit{Statistical Shape Models} (SSMs) address variability by aligning training shapes and learning a low-dimensional subspace for population-level morphometrics. They have been widely adopted for segmentation, measurement, and diagnosis~\cite{heimann2009statistical,ambellan2019statistical}, and can even generate patient-specific bone models from sparse 3D/2D data~\cite{schmid2011robust,baka2011shape}. Nevertheless, classical SSM construction depends on manual landmark annotation, explicit correspondence, and PCA, which limits scalability and automation on large medical datasets and can constrain the expressiveness of subtle anatomical details.

In parallel, modern segmentation networks---U-Net variants~\cite{ronneberger2015unet,chen2021transunet,ruan2024vmunet}, nnFormer~\cite{zhou2021nnformer}, and SAM-based extensions for medical imaging~\cite{kirillov2023segment,wang2023sammed3d,ma2024segment}---deliver robust voxel-level delineations. Yet a persistent gap remains between these voxel labels and clinically usable 3D models: meshes reconstructed from discrete masks are often jagged, topologically inconsistent, or over-smoothed, especially under limited resolution or incomplete coverage. This motivates learning shape priors directly from data so that sparse or noisy anatomical evidence can be lifted to dense, high-quality surfaces.

Recent studies are beginning to bridge point cloud learning and medical applications. Adams and Elhabian~\cite{adams2023can} explored unsupervised point-distribution modeling from raw point sets, while MedShapeNet~\cite{li2023medshapenet} established a large-scale benchmark for medical 3D shape learning. These advances indicate that \emph{implicit}, data-driven priors can complement or replace explicit SSMs in clinical 3D modeling, particularly when combined with strong segmentation front-ends.

\begin{figure*}[htbp]
\includegraphics[width=0.9\textwidth]{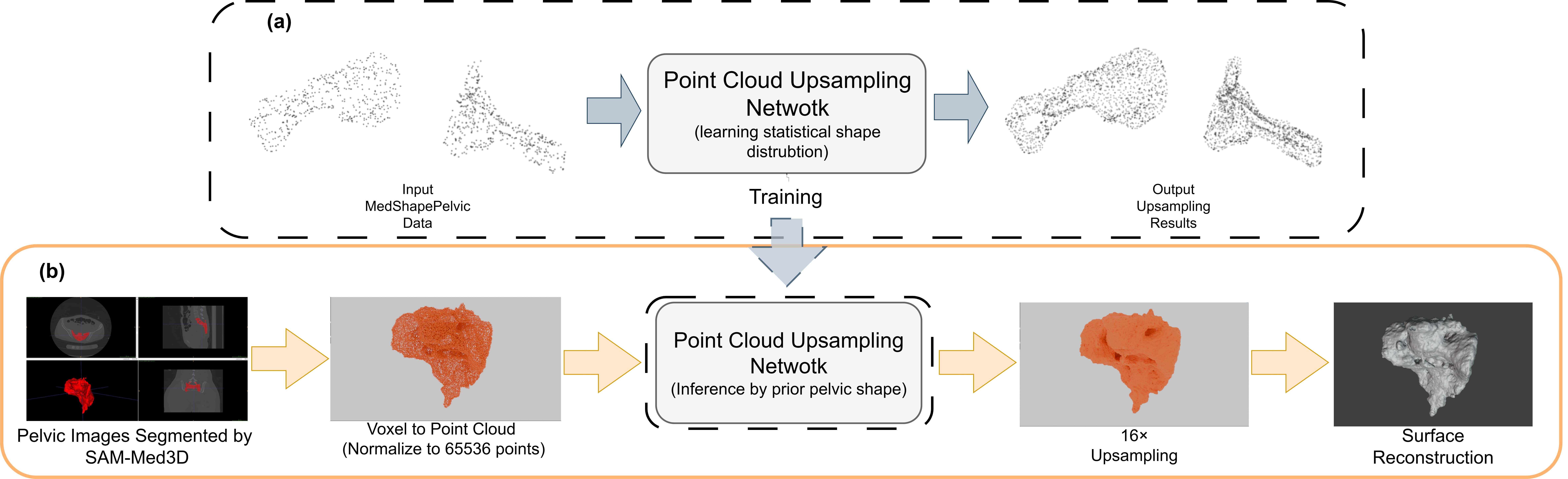}
\caption{Overview of the proposed pipeline. Medical images are first segmented into coarse anatomical masks (e.g., SAM-Med3D), then converted to point clouds, followed by upsampling to reconstruct a complete and high-resolution 3D bone model. Unlike classical Statistical Shape Models that require landmarks and PCA, our framework learns shape variability implicitly from dense point cloud data.}
\label{fig:1}
\end{figure*}

\paragraph{This work}
We propose a data-driven framework for producing clinically amenable meshes from segmentation outputs. Instead of explicitly constructing and sampling a statistical shape space, we leverage point cloud upsampling networks trained on large-scale 3D medical shape data to \emph{implicitly} encode anatomical priors. Using the pelvic subset of MedShapeNet (MedShapePelvic) as shape supervision and SAM-Med3D as a volumetric segmentation front-end, our framework learns to complete and refine sparse, segmentation-derived point sets into dense, anatomically faithful surfaces. This shift from explicit parametric modeling to learned priors enables flexible, automated reconstruction under varying input densities, helping to bridge the gap between voxel labels and high-quality meshes required in downstream clinical workflows.

\paragraph{Contributions}
\begin{itemize}
  \item \textbf{A segmentation-to-mesh pipeline with implicit priors.} We integrate SAM-Med3D-based voxel segmentation with deep point cloud upsampling to learn an implicit anatomical prior that densifies and regularizes segmentation-derived point sets, producing smooth, watertight meshes without manual landmarks or PCA.
  \item \textbf{Comprehensive evaluation on pelvic CT.} Trained on MedShapePelvic and validated on Pelvic1k, the proposed method is benchmarked against state-of-the-art upsampling approaches using geometry- and surface-level metrics (Chamfer/Hausdorff distances, point-to-surface error, F-score, normal consistency, and edge-aware measures), demonstrating consistent gains in surface quality and anatomical fidelity.
  \item \textbf{Generalization potential beyond the pelvis.} Although evaluated on pelvic bones, the framework is anatomy-agnostic and applicable to other skeletal regions and organs, providing a practical bridge from voxel segmentation to clinically usable 3D models in broader biomedical contexts.
\end{itemize}

\section{Related Works}

\subsection{Point Cloud Processing}
\subsubsection{In 3D Object}
Since the introduction of PointNet\cite{qi2017pointnet,qi2017pointnet++}, direct processing of raw point sets avoids the loss of context and complex steps involved in voxelization or multi-view conversion. Point cloud completion, in particular, is used to extract global features of 3D models, which can help supplement geometric details\cite{lin2024infocd,lin2024loss}. With the aid of deep learning and large-scale 3D datasets, learning-based methods have achieved excellent performance in shape completion tasks. For example, PCN\cite{yuan2018pcn} first employs an encoder to extract features and outputs dense and completed point clouds from sparse and incomplete input. Later, ASFM-Net\cite{xia2021asfmnet} used parallel global and local feature matching to reasonably infer missing geometric details of objects.

Besides, PU-Net\cite{yu2018punet} is a pioneering work that introduces CNNs for point cloud upsampling. Subsequently, PU-GCN\cite{qian2021pugcn} employed a graph-based network, achieving strong upsampling performance. Grad-PU\cite{he2023gradpu} provides flexible upsampling at different scales, while PUCRN\cite{du2022cascaded} optimized generation through multi-stage cascades. Transformer-based frameworks\cite{zhao2021pointtransformer} also improved robustness in sparse or missing regions. However, these methods are often trained on synthetic datasets (e.g., PU1k\cite{qian2021pugcn}) and may struggle to generalize to real anatomical shapes with irregular geometry and sparse sampling. Moreover, they typically focus on object-level CAD models and are not directly validated on medical segmentation pipelines.

\subsubsection{In Medical Image}
Recently, point cloud learning has been increasingly applied to medical image analysis. Chen et al.\cite{chen2021shape} proposed a learning-based deformable registration framework (MR-Net) to reconstruct 3D cardiac meshes from sparse 2D contours, showing real-time capability even with incomplete annotations. Beetz et al. developed \textit{Point2Mesh-Net}, which combines point-based encoders with mesh-based decoders for cardiac surface reconstruction from cine MRI slices\cite{beetz2022point2mesh}
, and later extended this line into multi-objective autoencoders for myocardial infarction prediction, enabling explainable outcome modeling from 3D cardiac shapes\cite{beetz2023multi}.
Hu et al.\cite{hu2021point}
introduced a GAN with tree-structured graph convolutions to generate 3D brain point clouds from a single 2D MRI slice. Jiang et al.\cite{jiang2023cpaconv}
proposed CPAConv-POCO, which leverages continuous position-adaptive convolution for lung nodule 3D reconstruction. Chen et al.\cite{chen2024cartilage}
applied completion networks to cartilage repair, using transformer-enhanced point cloud completion to recover femoral cartilage surfaces. 
For anatomical organ-level segmentation, Zhang et al.\cite{xzhang2023anatomical,xzhang2024robust}
designed point-voxel fusion networks to segment Couinaud liver segments, showing superior performance on LiTS and 3Dircadb. More recently, Yassin et al.\cite{yassin2024medshapenet} proposed a multimodal foundation model for medical point cloud completion based on the large-scale MedShapeNet dataset\cite{li2023medshapenet}, achieving generalizable completion across 240+ anatomical classes. Zhang et al.\cite{zhang2025hierarchical} introduced a hierarchical feature learning framework based on state-space models to further improve anatomical understanding tasks such as completion and segmentation. 
Although point cloud completion shares similarities with upsampling by extracting global features from partial input to complement geometric details, most existing methods are validated on synthetic or restricted datasets. In contrast, our approach explicitly integrates a medical image segmentation model (SAM-Med3D) with point cloud upsampling to form a complete pipeline from imaging to shape modeling. Furthermore, by training on realistic datasets such as MedShapePelvic and evaluating on the Pelvic1k subset, our method captures clinically relevant anatomical features and improves the quality of reconstructed surfaces.

\subsection{Segment Anything}
The Segment Anything Model (SAM)\cite{kirillov2023segment} was trained on a dataset of 11 million images and over 1 billion masks. This design allows the model to be enabled zero-shot transfer to new image distributions and tasks. Due to its speed and versatility, SAM's zero-shot capabilities have led to its use in medical image segmentation tasks, giving rise to MedSAM\cite{ma2024segment}. The MedSAM model\cite{ma2024segment} was developed on a large-scale medical image dataset containing 1.5 million images and masks, covering various imaging modalities and tumor types. It offers better accuracy and robustness than modality-specific models. However, MedSAM\cite{ma2024segment} only trained by organs and skin, including endoscopic images, and lacks 3D medical images, because they focus on interactive in 2D images. Later, SAM-Med3D\cite{wang2023sammed3d} was introduced, focusing on volumetric medical datasets for training and producing satisfactory 3D results. Notably, its training targets include not only organs, skin, and cells but also bones. It is worth mentioning that SSM represents shape variation through statistical analysis of anatomical variations in shape populations generated from medical images, and it has applications in multiple medical fields, including facial recognition and skull analysis\cite{cai2024statistical}. This drives our objective of combining deep learning methods for medical image segmentation and point cloud upsampling, enabling SSM modeling and shape variation.

Ergo, our study leverages SAM-Med3D as a frontend for downstream statistical shape modeling. By combining its segmentation output with point cloud upsampling, we demonstrate that deep segmentation models can serve as a viable input source for high-fidelity shape reconstruction.

\section{Methodology}

\subsection{Overview of the Proposed Framework}



We propose a novel framework that integrates medical image segmentation and point cloud upsampling for high-fidelity anatomical reconstruction. As shown in Figure~\ref{fig:1}(a), the core idea is to learn an implicit anatomical prior from real data through a deep upsampling network, allowing sparse point clouds to be restored to dense and anatomically plausible representations.

The pipeline consists of two sequential modules. First, we apply the SAM-Med3D model to volumetric CT data to obtain voxel-wise segmentation masks. These masks are converted into sparse point clouds using spatial coordinates. Then, as illustrated in Figure~\ref{fig:1}(b), the sparse point cloud is passed into a pretrained point cloud upsampling network, which has been trained on MedShapePelvic samples. Finally, surface reconstruction is applied to the dense point cloud using Marching Cubes, yielding smooth triangular meshes suitable for quantitative evaluation and visualization.

\begin{figure}[htbp]
\centerline{\includegraphics[width=0.5\textwidth]{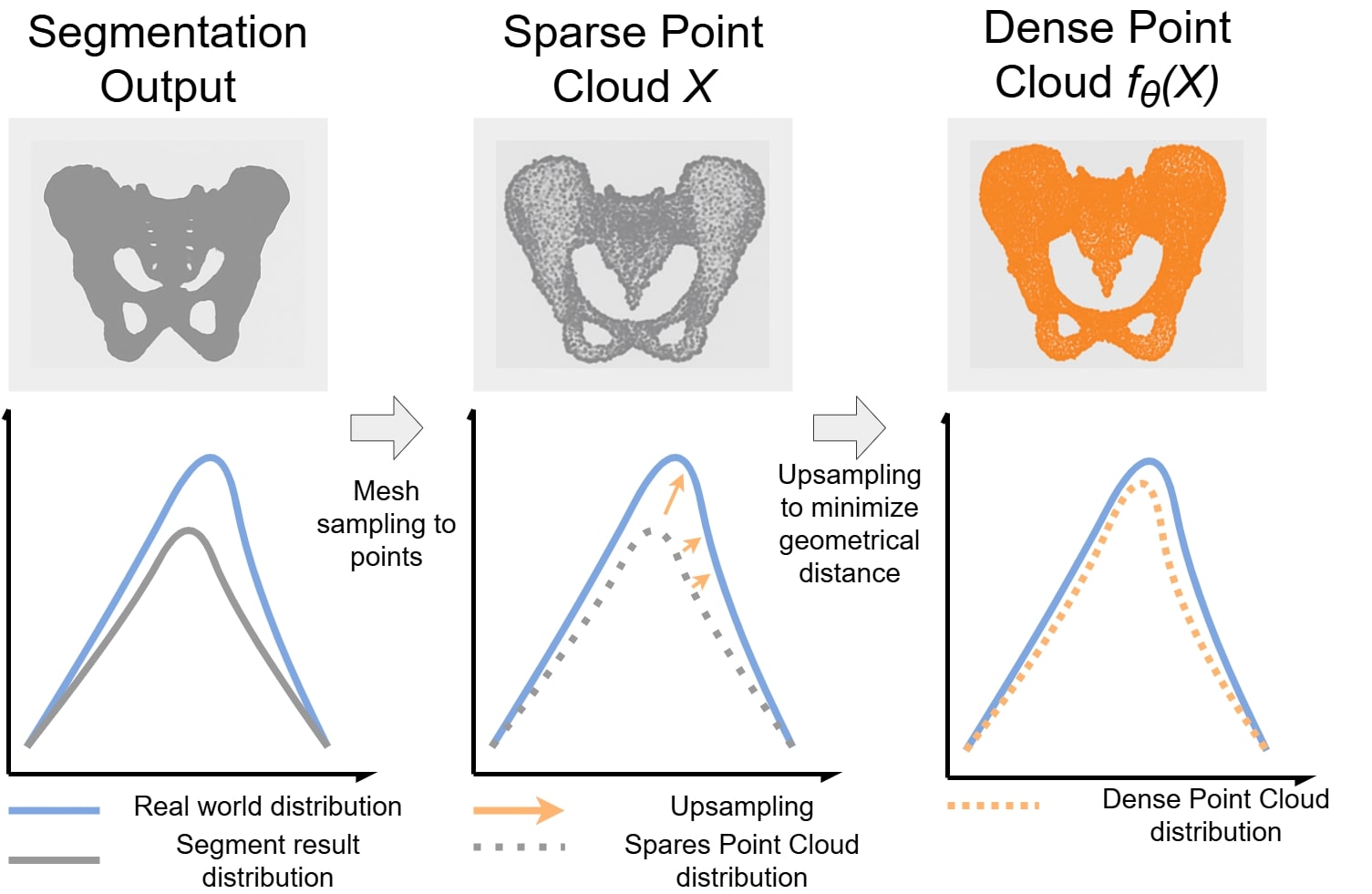}}
\caption{Distributional interpretation of implicit anatomical prioring. The segmentation-derived point cloud (left) deviates from the real anatomical distribution. Our network $f_\theta$ learns to map the sparse inputs toward the ground-truth distribution through point cloud upsampling. This reflects an implicit learning of intra-population variation, capturing not only shape completion but also a shift toward the anatomical manifold.}
\label{fig:math_alignment}
\end{figure}

\subsection{Learning Implicit Anatomical Priors}

\paragraph{Distributional notation}
Let $X=\{x_i\}_{i=1}^{N}\subset\mathbb{R}^3$ be the sparse points sampled from SAM\mbox{-}Med3D masks and $Y=\{y_j\}_{j=1}^{M}$ the dense target points from MedShapePelvic.
We regard point clouds as empirical measures
$\mu_X=\frac{1}{N}\sum_{i=1}^{N}\delta_{x_i}$ and $\nu_Y=\frac{1}{M}\sum_{j=1}^{M}\delta_{y_j}$.
The upsampler $f_\theta:\mathbb{R}^{N\times 3}\!\to\!\mathbb{R}^{M\times 3}$ induces a pushforward measure
$f_{\theta:}\mu_X=\frac{1}{M}\sum_{j=1}^{M}\delta_{\hat y_j}$ with $\hat Y=f_\theta(X)=\{\hat y_j\}_{j=1}^M$.
Learning then amounts to aligning $f_{\theta:}\mu_X$ with $\nu_Y$ on the anatomical shape manifold approximated by the training set.

\paragraph{Empirical risk}
We minimize a symmetric Chamfer discrepancy between the predicted and target point sets: Eq.~\eqref{eq:loss_chamfer_measure}

\begin{figure*}[htbp]
\centering
\begin{equation}
\label{eq:loss_chamfer_measure}
\min_{\theta}\;
\mathbb{E}_{(X,Y)\sim\mathcal{D}}
\Big[
\underbrace{
\int \!\! \min_{y\in Y}\|z-y\|_2^2\, \mathrm{d}\big(f_{\theta:}\mu_X\big)(z)
}_{\text{pred}\to\text{target}}
\;+\;
\underbrace{
\int \!\! \min_{\hat y\in \hat Y}\|z-\hat y\|_2^2\, \mathrm{d}\nu_Y(z)
}_{\text{target}\to\text{pred}}
\Big],
\end{equation}
\end{figure*}

which is exactly the usual (squared) Chamfer distance written in a distributional form.
In practice, Eq.~\eqref{eq:loss_chamfer_measure} reduces to
\[
\mathrm{CD}(\hat Y,Y)
=\frac{1}{M}\sum_{\hat y\in \hat Y}\min_{y\in Y}\|\hat y-y\|_2^2
+\frac{1}{M}\sum_{y\in Y}\min_{\hat y\in \hat Y}\|y-\hat y\|_2^2 .
\]
Optimizing Eq.~\eqref{eq:loss_chamfer_measure} embeds shape regularity and population variability into the network parameters $\theta$, enabling robust densification of sparse, segmentation-derived inputs without landmarks, explicit correspondences, or PCA.

Figure~\ref{fig:math_alignment} further illustrates this concept from a distributional perspective, showing how our model maps the segmentation-derived distribution toward the ground-truth anatomical distribution via upsampling. We observe a consistent reduction in Chamfer Distance variance across test samples after upsampling, indicating convergence to a learned anatomical distribution.

Ergo, the network aims to reduce geometric variance and improving reconstruction fidelity across test cases.

\subsection{Evaluation Metrics}

\subsubsection{For Point Cloud Upsampling}
To evaluate the point cloud upsampling network, we used three key metrics: 


\textbf{Chamfer Distance (CD)} measures the average of the nearest-neighbor distances between two point sets. Let \( P \) and \( Q \) be the two sets of points (the upsampled point cloud and the ground truth point cloud, respectively), the Chamfer Distance is defined as:

\[
\text{CD}(P, Q) = \frac{1}{|P|} \sum_{p \in P} \min_{q \in Q} \| p - q \|^2 + \frac{1}{|Q|} \sum_{q \in Q} \min_{p \in P} \| q - p \|^2
\]


\textbf{Hausdorff Distance (HD)} captures the maximum distance between points in one set and their closest point in the other set, thus highlighting the worst-case mismatch. The Hausdorff distance between sets \( P \) and \( Q \) is defined as:

\[
\text{HD}(P, Q) = \max \left( \max_{p \in P} \min_{q \in Q} \| p - q \|, \max_{q \in Q} \min_{p \in P} \| q - p \| \right)
\]


\textbf{Point-to-Surface Distance (P2F)} evaluates the shortest distance from upsampled points to the surface of the reference model. Given a set of upsampled points \( P \) and a reference surface \( S \), the Point-to-Surface Distance is defined as:

\[
\text{P2F}(P, S) = \frac{1}{|P|} \sum_{p \in P} \min_{s \in S} \| p - s \|
\]


\subsubsection{For Surface Reconstruction}\label{sec:3.4.2}
To evaluate the surface reconstruction of our framework, we used three key metrics: Area-Length Ratio, Manifoldness Rate and Connected Component Discrepancy\cite{guo2024tetsphere,botsch2010polygon}. 

\textbf{Area-Length Ratio (ALR)} measures the shape quality of individual triangles within a mesh. This ratio attains its maximum value of 1 for equilateral triangles, indicating optimal shape quality.

\textbf{Manifoldness Rate (MR)} quantifies the proportion of edges in a mesh that satisfy manifold conditions. An edge is considered manifold if it is shared by exactly two faces. A rate approaching 1 indicates a well-structured mesh suitable for operations like Boolean computations and 3D printing. Conversely, a lower rate highlights the presence of non-manifold edges, which may necessitate mesh repair or refinement.

\textbf{Connected Component Discrepancy (CC Diff.)} assesses the topological connectedness of a mesh by comparing the observed number of connected components to the expected number. A discrepancy of 0 signifies that the mesh has the anticipated number of connected components, implying structural integrity.

Those three indicators to evaluate the geometric quality of reconstructed meshes.

Besides, to evaluate the fidelity of the reconstructed meshes against the ground truth, we employ several widely recognized metrics\cite{chen2022ndc}: \textbf{Chamfer Distance (CD)}, \textbf{F-Score (F1)}, \textbf{Normal Consistency (NC)}, \textbf{Edge Chamfer Distance (ECD)}, and \textbf{Edge F-Score (EF1)}.

\begin{table*}[htbp]
\caption{Performance comparison of two methods with MedShapePelvic across input scales.}
\centering
\resizebox{0.8\textwidth}{!}{%
\begin{tabular}{ccccccccccccc}
\hline
\multirow{2}{*}{Method} & \multicolumn{3}{c}{512 input}              & \multicolumn{3}{c}{1024 input}             & \multicolumn{3}{c}{2048 input}             & \multicolumn{3}{c}{4096 input}             \\ \cline{2-13} 
                         & CD$\downarrow$ & HD$\downarrow$ & P2F$\downarrow$ & CD$\downarrow$ & HD$\downarrow$ & P2F$\downarrow$ & CD$\downarrow$ & HD$\downarrow$ & P2F$\downarrow$ & CD$\downarrow$ & HD$\downarrow$ & P2F$\downarrow$ \\ \hline
MPU               & 13.050         & 11.971        & 10.604          & 8.507          & 26.088        & 8.759           & 5.718          & 9.421        & 8.558          & 3.906          & 8.768        & 7.556           \\
PU-GCN            & 4.191          & 9.942          & 8.553           & 5.326          & 21.284         & 7.329           & 2.483          & 7.296          & 4.652           & 1.719          & 6.519          & 3.671           \\
Grad-PU                  & 5.643          & 124.639        & 12.438          & \textbf{3.682}          & 82.969         & 9.779           & 2.477          & 49.451         & 4.852           & 1.690          & 28.871         & 3.172           \\
PUCRN                    & \textbf{3.104}          & \textbf{7.806}          & \textbf{6.150}           & 3.946          & \textbf{16.635}         & \textbf{6.436}           & \textbf{1.839}          & \textbf{5.709}          & \textbf{4.066}           & \textbf{1.274}          & \textbf{5.087}          & \textbf{2.719}           \\ \hline
\end{tabular}%
}
\label{tab:1}
\end{table*}

\section{Experiments}

\subsection{Settings}

\subsubsection{Datasets}
In our study, experimental validation on the pelvic region have been conducted in the hips, femurs, and sacrum from MedShapeNet\cite{li2023medshapenet} as MedShapePelvic for point cloud upsampling to learn shape prior knowledge. We then visualized Quantitative analyzed result based on Pelvic1k\cite{liu2021deep}, a comprehensive pelvic CT dataset capable of replicating actual appearance changes. Due to the lack of maintenance in some of the open source data used by Pelvic1k, we can only obtain the following data (e.g. subdatasets 3, 4, 6).

For training and testing the upsampling network, we used the MedShapePelvic dataset based on MedShapeNet\cite{li2023medshapenet}, which includes 50,496 training samples from 3,156 Mesh models generated via patch sampling\cite{yifan2019patch, qian2021pugcn,zhang2024rethinking,ZHANG2025104467}. The dataset consists of 3,506 3D models, split into 3,156 training samples and 350 testing samples. Patch sampling was performed by generating training data through Poisson disk sampling on patches of the 3D meshes.

\subsubsection{Comparison Methods}
In order to verify the effectiveness of our proposed theory, point cloud upsampling is the first part. We built a baseline based on MPU\cite{yifan2019patch}, PU-GCN\cite{qian2021pugcn}, Grad-Pu\cite{he2023gradpu} and PUCRN\cite{du2022cascaded}. In more experiments, only by comparing it with two advanced algorithms: Grad-Pu that realized any ratio upsampling and PUCRN that is the SOTA model.

For a fair and objective comparison, we obtained the open-source codes of these methods and trained and tested them on our computing equipment. And we test the result based on training in MedShapePelvic\cite{li2023medshapenet} and PU1k dataset\cite{qian2021pugcn}, which is a synthetic dataset from 3D models. This comparison with identical parameter settings would be presented in ablation: Section \ref{sec:4}.

\subsubsection{Implementation Details}
Experiments were developed using the PyTorch framework on the Ubuntu 22.04 system. We used an NVIDIA L20 GPU with 48GB of graphics memory and an Intel(R) Xeon(R) Platinum 8457C host with 100GB of RAM. The network was trained over 100 epochs with a batch size of 64. The initial learning rate was set to 0.001, with a decay rate of 0.05.

\begin{table}[htbp]
\caption{Quantitative comparison in $16\times$ upsampling after training in $4\times$. The input points are 512 and 1024.}
\centering
\begin{tabular}{ccccccc}
\hline
\multirow{2}{*}{Method} & \multicolumn{3}{c}{512 to 8192}                   & \multicolumn{3}{c}{1024 to 16384}                 \\ \cline{2-7} 
                         & CD$\downarrow$ & HD$\downarrow$ & P2F$\downarrow$ & CD$\downarrow$ & HD$\downarrow$ & P2F$\downarrow$ \\ \hline
Grad-PU                  & 5.827          & 124.640        & 13.117          & 3.422          & 84.825         & 10.329          \\
PUCRN                    & 3.586          & 9.984          & 6.824           & 2.870          & 18.899         & 5.872           \\ \hline        
\end{tabular}
\label{tab:2}
\end{table}

\begin{table*}[htbp]
\caption{Pelvic point cloud upsampling results in two different train sets: \textbf{Network(trainset)}.}
\centering
\resizebox{0.8\textwidth}{!}{%
\begin{tabular}{ccccccccc}
\hline
\multirow{2}{*}{Method} & \multicolumn{2}{c}{512 input} & \multicolumn{2}{c}{1024 input} & \multicolumn{2}{c}{2048 input} & \multicolumn{2}{c}{4096 input} \\ \cline{2-9} 
                         & CD$\downarrow$    & HD$\downarrow$   & CD$\downarrow$    & HD$\downarrow$    & CD$\downarrow$    & HD$\downarrow$    & CD$\downarrow$    & HD$\downarrow$    \\ \hline
Grad-PU(MedShapePelvic)     & 5.643             & 124.639          & 3.682             & 82.969            & 2.477             & 49.451            & 1.690             & 28.871            \\
PUCRN(MedShapePelvic)       & \textbf{3.104}    & \textbf{7.806}   & 3.946             & 16.635            & \textbf{1.839}    & \textbf{5.709}    & \textbf{1.274}    & \textbf{5.087}    \\
Grad-PU(PU1k)            & 5.411             & 15.551           & 3.353             & 12.280            & 2.256             & 8.677             & 1.513             & 7.872             \\
PUCRN(PU1k)              & 3.749             & 10.065           & \textbf{2.902}    & \textbf{9.955}    & 1.995             & 6.134             & 1.379             & 5.253             \\ \hline
\end{tabular}%
}
\label{tab:3}
\end{table*}

\subsection{Upsampling Results}
In accuracy analysis, as mentioned above, the generalization performance of two SOTA network have been evaluated on the MedShapePelvic dataset in $4\times$ upsampling. The quantitative results for the dataset in Table \ref{tab:1} show that our method performs consistently across different input scales (512, 1024, 2,048 and 4,096 points) and outperforms the original network in CD, HD, and P2F metrics. 

This result demonstrates the effectiveness of PUCRN in CD, HD, and P2F, showcasing its superior capability in learning the shape prior knowledge in pelvic. The 3 stage network refinement structure allow it could minimizing nearest-neighbor distance differences and reducing maximum point-to-point distance errors. 

\begin{figure*}[htbp]
\centering
	
	\begin{minipage}{0.8\linewidth}
		\centerline{\includegraphics[width=0.9\textwidth]{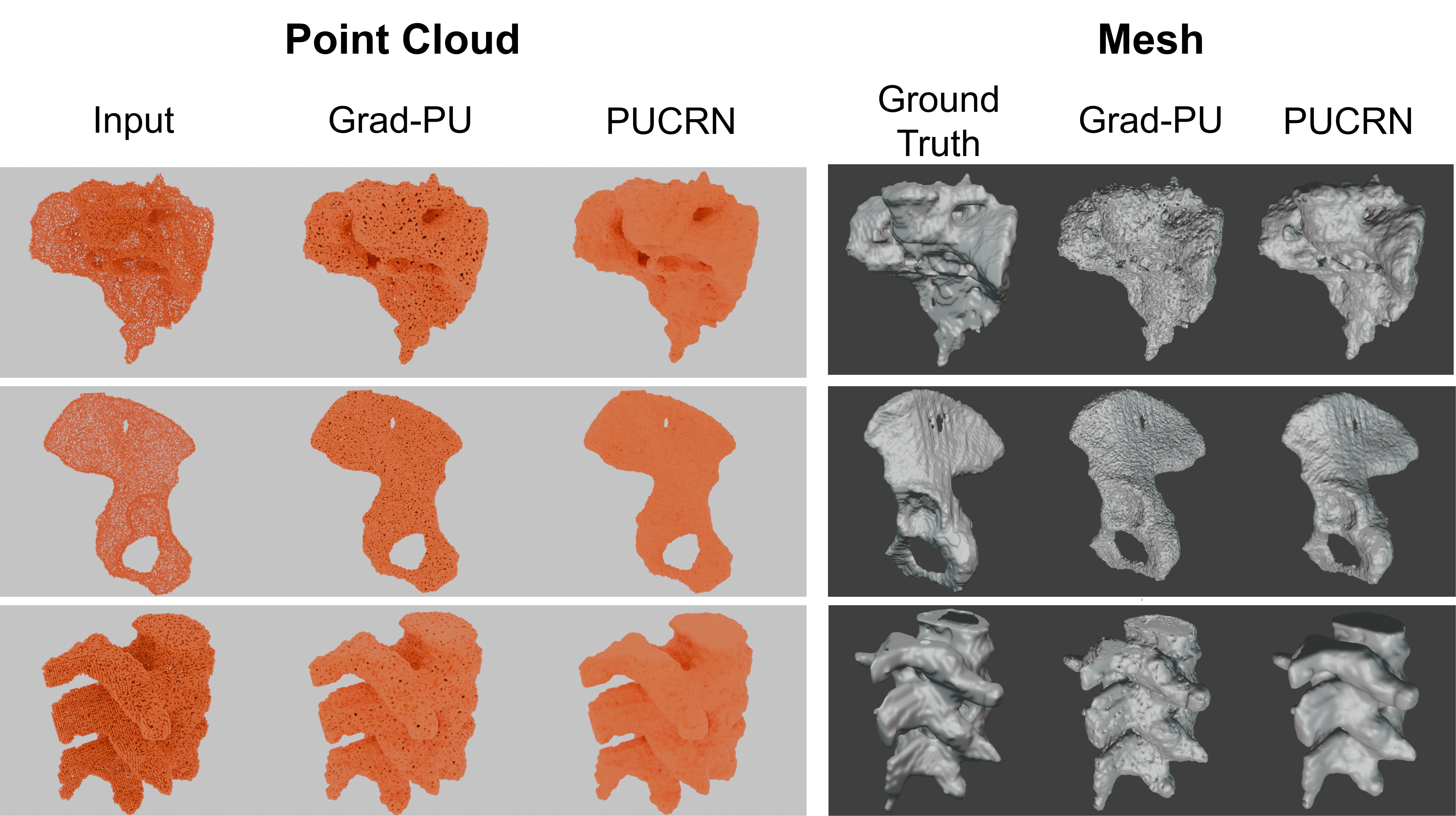}}
		\centerline{\includegraphics[width=0.9\textwidth]{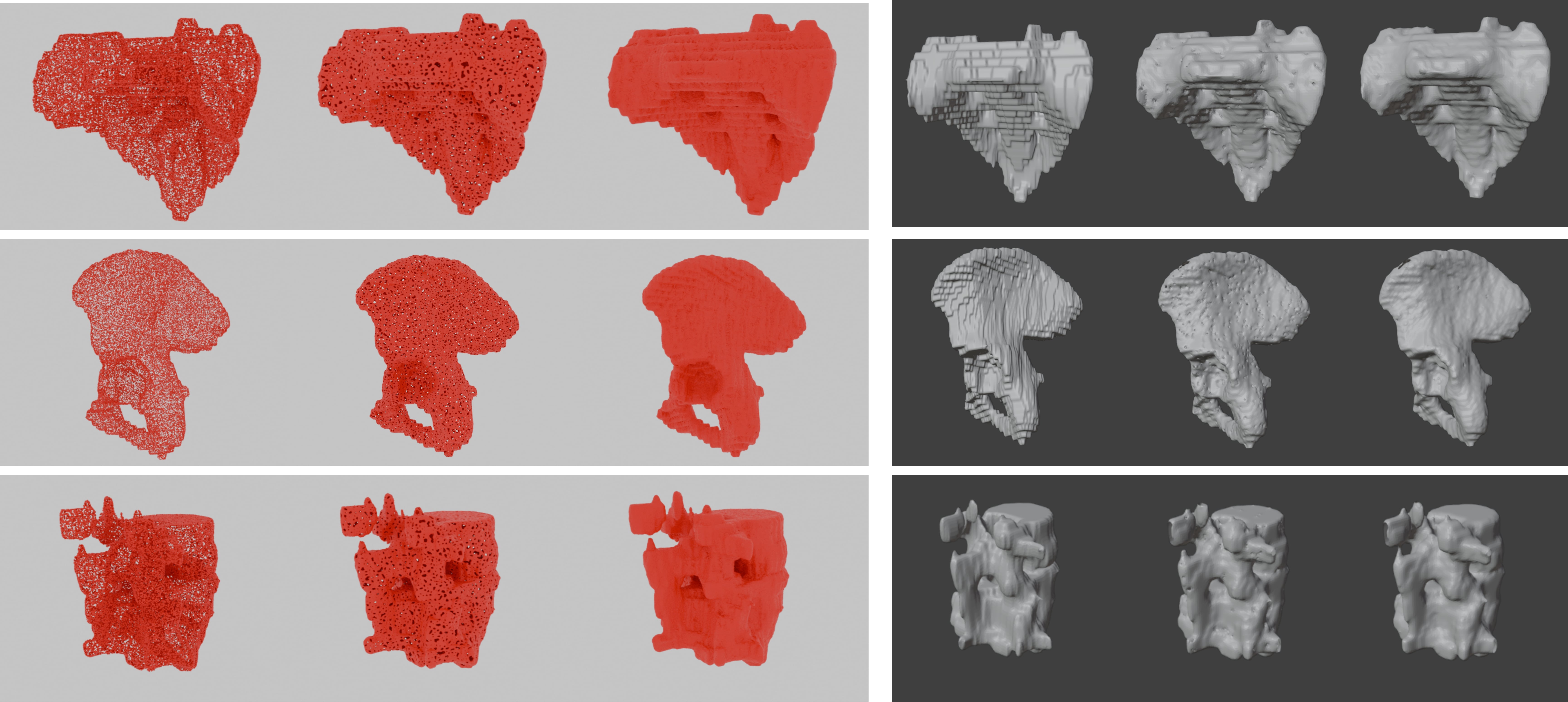}}
	\end{minipage}
	 
\caption{Point cloud upsampling results on Pelvic1k segment sparse inputs. Compared with the voxel surface in segmented results, our method can generate more uniform with detailed structures.} \label{fig:2}
\end{figure*}

Additionally, noticed that Grad-PU with very high HD. This may be caused by the input pelvic training data. This reflects that Grad-PU, lacking local and global refinement, cannot achieve geometric differences based on Pelvic training data, in Figure \ref{fig:2} showed that the upsampling results of Grad-PU have many holes which means it cannot learn good enough in local features. 

Besides, we also conducted $16\times$ upsampling tests, in Table \ref{tab:2}. Grad-PU can perform upsampling at any scale, so we set its ratio directly to $16\times$ on the test set. However, the testing model of PUCRN can only use the same upsampling factor as the training conditions, so we performed two $4\times$ samplings on PUCRN to achieve a $16\times$ sampling effect. The results show that PUCRN still has good performance.

\subsection{Ablation Studies} \label{sec:4}
To verify the effectiveness and necessity of training in MedShapePelvic to learn the prior knowledge of shape, the $4\times$ sampling result based on training in MedShapePelvic and PU1k dataset\cite{qian2021pugcn} have been tested. Among them, MedShapePelvic has 50496 training samples and PU1k has 69000 training samples. This avoids the difference in learning effectiveness caused by the large difference in the number of training samples. The result shows in Table \ref{tab:3}

As a result, training on the MedShapePelvic dataset performed better than training on PU1k. This indicates that PUCRN has indeed learned Pelvic's shape features, which means that it learned statistical variation embedded in anatomical structures, serving as an implicit anatomical prior without relying on PCA or landmarks. However, Grad-PU on the PU1k dataset performed better than MedShapePelvic's test set. These reasons are multifaceted, and Grad-PU may be better able to learn shape features and knowledge from artificially synthesized PU1k datasets, rather than being adept at learning features from MedShapePelvic obtained from real volume medical images.

\begin{table*}[htbp]
\caption{Surface reconstruction results on Pelvic1k subdataset by mesh quality metrics: Area-Length Ratio (ALR), Manifoldness Rate (MR), and Connected Component Discrepancy (CC Diff.).}
\centering
\resizebox{0.8\textwidth}{!}{%
\begin{tabular}{cccccccccc}
\hline
                                & \multicolumn{3}{c}{Subdataset 3}                                                                   & \multicolumn{3}{c}{Subdataset 4}                                                                   & \multicolumn{3}{c}{Subdataset 6}                                                                   \\ \cline{2-10} 
\multirow{-2}{*}{Mesh}          & ALR$\uparrow$ & MR(\%)$\uparrow$ & CC Diff.$\downarrow$ & ALR$\uparrow$ & MR(\%)$\uparrow$ & CC Diff.$\downarrow$ & ALR$\uparrow$ & MR(\%)$\uparrow$ & CC Diff.$\downarrow$ \\ \hline
Ground Truth                    & 0.692                       & 100.0                          & 0.26                             & 0.721                       & 100.0                          & 1.68                             & 0.742                       & 100.0                          & 0.66                             \\
Segmented Result (by SAM-Med3D) & 0.676                       & 100.0                          & 1.94                             & 0.699                       & 100.0                          & 4.89                             & 0.735                       & 100.0                          & 1.18                             \\
Med-PU (by Grad-PU)              & 0.811                       & 100.0                          & 2.77                             & \textbf{0.817}                       & 99.9                           & 9.66                             & 0.819                       & 100.0                          & 7.77                             \\
Med-PU (by PUCRN)                & \textbf{0.821}                       & 100.0                          & 2.06                             & \textbf{0.817}                       & 100.0                          & 5.72                             & \textbf{0.824}                       & 100.0                          & 5.34                             \\ \hline
\end{tabular}%
}
\label{tab:4}
\end{table*}

\subsection{Reconstruction Visualizations}
In addition to the results of point cloud upsampling, what we need to achieve is the combination of medical image segmentation and point cloud upsampling steps to achieve the effect of SSM. 
\begin{table*}[htbp]
\caption{Surface reconstruction quantitative results on on Pelvic1k subdataset with 16,384 points by poisson sampling.}
\centering
\resizebox{0.9\textwidth}{!}{%
\begin{tabular}{cccccccccccccccc}
\hline
                                                                                                   & \multicolumn{5}{c}{Subdataset 3}                                                                           & \multicolumn{5}{c}{Subdataset 4}                                                                           & \multicolumn{5}{c}{Subdataset 6}                                                                           \\ \cline{2-16} 
\multirow{-2}{*}{\begin{tabular}[c]{@{}c@{}}16,384 Points\\ by Poisson Sampling\end{tabular}} & CD$\downarrow$     & F1$\uparrow$    & NC$\uparrow$ & ECD$\downarrow$ & EF1$\uparrow$ & CD$\downarrow$     & F1$\uparrow$    & NC$\uparrow$ & ECD$\downarrow$ & EF1$\uparrow$ & CD$\downarrow$     & F1$\uparrow$    & NC$\uparrow$ & ECD$\downarrow$ & EF1$\uparrow$ \\ \hline
Segmented Result (by SAM-Med3D)                                                                    & 8.379                  & 0.214                  & 0.691                                          & 7.420                                           & 0.408                                           & 56.272                 & \textbf{0.126}                  & \textbf{0.711}                                          & 54.559                                          & 0.250                                           & 13.997                 & 0.214                  & \textbf{0.716}                                          & 10.724                                          & 0.357                                           \\
Med-PU (by Grad-PU)                                                                                 & 7.200                  & 0.223                  & 0.671                                          & 7.326                                           & 0.445                                           & 53.297                 & 0.109                  & 0.651                                          & 50.104                                          & 0.254                                           & 12.024                 & 0.223                  & 0.658                                          & 8.351                                           & 0.453                                           \\
Med-PU (by PUCRN)                                                                                   & \textbf{7.111}                  & \textbf{0.229}                  & \textbf{0.766}                                          & \textbf{6.295}                                           & \textbf{0.464}                                           & \textbf{51.799}                 & 0.118                  & 0.678                                          & \textbf{49.824}                                          & \textbf{0.266}                                           & \textbf{11.620}                 & \textbf{0.229}                  & 0.686                                          & \textbf{8.184}                                           & \textbf{0.476}                                           \\ \hline
\end{tabular}%
}
\label{tab:5}
\end{table*}


The Pelvic1k subdataset 3,4 and 6 have been inputted into SAM-Med3D, and after medical image segmentation, we obtain predicted masks of the pelvic bone, which are then stacked to obtain voxels and converted into point clouds. The current point cloud is sparse, and the segmentation point cloud we obtained is standardized to input 65,536 points into the upsampling model and perform $16\times$ upsampling. The model surface is reconstructed using the \textbf{Voxel Grid with Marching Cubes}\cite{lorensen1998marching}, where voxel size is $1.5$. As shows in Figure \ref{fig:2}, compared with the surface of ground truth, and obtained by directly segmenting voxels through marching cubes. Our proposed framework achieves smoother after upsampling in both Grad-PU and PUCRN and more complete shape reconstruction in PUCRN. And there are no obvious contours of layer after reconstructed.

\subsection{Reconstruction Evaluations}
To evaluate the quality of the mesh surface vis-a-vis ground truth, we used three metrics in Section\ref{sec:3.4.2} \textbf{For Surface Reconstruction}.

As shows in Table \ref{tab:4}. Our method outperforms ground truth and segmented result in terms of mesh quality and achieves stable reconstruction accuracy. Specifically, our method has uniform grid partitioning (as shown in ALR) and eliminates floating artifacts (as shown in CC Diff.). This improvement is attributed to the finer points and distribution brought about by point cloud upsampling.

Furthermore, we conducted quantitative comparisons on the reconstruction results in order to evaluate surface reconstruction accuracy and feature preservation. We used metrics (CD, F1, NC, ECD, EF1) in Section\ref{sec:3.4.2} \textbf{For Surface Reconstruction}.

To ensure a fair and uniform comparison, we employ Poisson disk sampling to generate 16,384 points from each mesh, including the ground truth, segmented results, and upsampled reconstructions. This sampling strategy maintains uniform point distribution and adequately captures both global structures and fine details, facilitating reliable metric computation.

As shows in Table \ref{tab:5}. Our method, across all three subdatasets, PUCRN consistently outperforms both Grad-PU and the raw SAM-Med3D segmentations. For example, on Subdataset 3, PUCRN achieves a significantly lower CD compared to Grad-PU, along with superior edge preservation as reflected by a lower ECD and a higher EF1-score. Similar trends are observed on Subdataset 4 and Subdataset 6, where PUCRN achieves the best overall performance in terms of both global surface quality and local edge fidelity. These results demonstrate the effectiveness of PUCRN in enhancing the downstream surface reconstruction quality by providing denser and more accurate point distributions. And we observe consistent reductions in Chamfer Distance and improvements in Normal Consistency and F-scores, indicating that the upsampling network effectively maps sparse point clouds toward the anatomical shape manifold as Figure \ref{fig:math_alignment}.

\section{Discussion}
This study demonstrates that coupling a robust volumetric segmenter with a dedicated point cloud upsampler can bridge the gap between voxel labels and mesh-quality anatomical models. Compared with classical SSMs that rely on landmarks and explicit statistical spaces, our approach learns an \emph{implicit} prior from realistic pelvic shapes (MedShapePelvic), enabling automatic densification and regularization of sparse, segmentation-derived points into smooth, watertight surfaces suited for visualization and downstream analysis.

\paragraph{Why domain supervision matters}
Our ablations indicate consistent gains when the upsampler is trained on anatomical meshes (MedShapePelvic) rather than synthetic/CAD-style objects (PU1k). This aligns with the intuition that the statistics of cortical thickness, local curvature, and trabecular recesses in pelvic bones differ markedly from generic object datasets. Prior CMPB studies similarly show that medical-domain supervision/priors improve realism and downstream usability of the reconstructed geometry~\cite{cheng2020morphing,xi2021recovering}. In practice, domain-aligned supervision helps suppress perforations and over-smoothing, yielding lower HD and improved edge fidelity after meshing.

\paragraph{Limitations and Future Work}
The performance of our 3D reconstruction pipeline, Med-PU, is limited by several factors. First, it depends on the quality of front-end segmentation and is sensitive to imaging domain shifts, such as challenging boundaries or metal artifacts, which may propagate errors to the point set and surface. Second, our meshing relies on iso-surface extraction without explicit topology control, and while we apply standard cleanup and report manifoldness-oriented metrics, pathological non-manifold cases may still occur. Third, our evaluation is geometry-centric, and the clinical impact—such as interobserver variability in contouring, planning accuracy, or dose/navigation surrogates—remains to be quantified through future user studies. Broader evaluations across diverse anatomies, scanners, and acquisition protocols are needed to assess robustness to domain shifts. To address inherited segmentation errors, joint or end-to-end training of segmentation and upsampling could be explored. Additionally, runtime and memory demands increase with higher upsampling ratios, suggesting the need for lighter backbones, GPU-friendly operators, or uncertainty-aware pruning to enhance efficiency and reliability. Finally, integrating task-level endpoints and extending the implicit prior beyond medical applications (e.g., to robotics or industrial inspection) would further validate Med-PU as a general-purpose, knowledge-based 3D reconstruction system. Broader surveys, such as those in clinical 3D reconstruction for endoscopy, emphasize the importance of task-driven validation beyond geometric scores~\cite{yang20243d}.

\section{Conclusion}
In this work, we proposed Med-PU, an implicit anatomical prioring framework that leverages deep point cloud upsampling to reconstruct high-resolution pelvic structures from sparse medical data. Our method avoids the limitations of classical SSMs by eliminating the need for landmark annotations and explicitly modeling shape variation. Experiments demonstrate improved surface quality and anatomical plausibility across multiple evaluation metrics.

Instead of explicitly decomposing shape variations, Med-PU learns to embed these variations in a function space. This allows the model to restore anatomically plausible shapes from sparse or noisy segmentations, demonstrating its ability to capture and regularize shape diversity.

\bibliographystyle{IEEEtran}
\bibliography{reference}

\begin{thebibliography}{10}
\providecommand{\url}[1]{#1}
\csname url@samestyle\endcsname
\providecommand{\newblock}{\relax}
\providecommand{\bibinfo}[2]{#2}
\providecommand{\BIBentrySTDinterwordspacing}{\spaceskip=0pt\relax}
\providecommand{\BIBentryALTinterwordstretchfactor}{4}
\providecommand{\BIBentryALTinterwordspacing}{\spaceskip=\fontdimen2\font plus
\BIBentryALTinterwordstretchfactor\fontdimen3\font minus \fontdimen4\font\relax}
\providecommand{\BIBforeignlanguage}[2]{{%
\expandafter\ifx\csname l@#1\endcsname\relax
\typeout{** WARNING: IEEEtran.bst: No hyphenation pattern has been}%
\typeout{** loaded for the language `#1'. Using the pattern for}%
\typeout{** the default language instead.}%
\else
\language=\csname l@#1\endcsname
\fi
#2}}
\providecommand{\BIBdecl}{\relax}
\BIBdecl

\bibitem{heimann2009statistical}
T.~Heimann and H.~P. Meinzer, ``Statistical shape models for 3d medical image segmentation: a review,'' \emph{Medical image analysis}, vol.~13, no.~4, pp. 543--563, 2009.

\bibitem{ambellan2019statistical}
F.~Ambellan, H.~Lamecker, C.~von Tycowicz, and S.~Zachow, \emph{Statistical shape models: understanding and mastering variation in anatomy}.\hskip 1em plus 0.5em minus 0.4em\relax Springer International Publishing, 2019.

\bibitem{schmid2011robust}
J.~Schmid, J.~Kim, and N.~Magnenat-Thalmann, ``Robust statistical shape models for mri bone segmentation in presence of small field of view,'' \emph{Medical image analysis}, vol.~15, no.~1, pp. 155--168, 2011.

\bibitem{baka2011shape}
N.~Baka, B.~L. Kaptein, M.~de~Bruijne, T.~van Walsum, J.~E. Giphart, W.~J. Niessen, and B.~P. Lelieveldt, ``2d–3d shape reconstruction of the distal femur from stereo x-ray imaging using statistical shape models,'' \emph{Medical image analysis}, vol.~15, no.~6, pp. 840--850, 2011.

\bibitem{ronneberger2015unet}
O.~Ronneberger, P.~Fischer, and T.~Brox, ``U-net: Convolutional networks for biomedical image segmentation,'' \emph{Medical image computing and computer-assisted intervention--MICCAI 2015: 18th international conference, Munich, Germany, October 5-9, 2015, proceedings, part III}, pp. 234--241, 2015.

\bibitem{chen2021transunet}
J.~Chen, Y.~Lu, Q.~Yu, X.~Luo, E.~Adeli, Y.~Wang, and Y.~Zhou, ``Transunet: Transformers make strong encoders for medical image segmentation,'' \emph{arXiv preprint arXiv:2102.04306}, 2021.

\bibitem{ruan2024vmunet}
J.~Ruan and S.~Xiang, ``Vm-unet: Vision mamba unet for medical image segmentation,'' \emph{arXiv preprint arXiv:2402.02491}, 2024.

\bibitem{zhou2021nnformer}
H.~Y. Zhou, J.~Guo, Y.~Zhang, L.~Yu, L.~Wang, and Y.~Yu, ``nnformer: Interleaved transformer for volumetric segmentation,'' \emph{arXiv preprint arXiv:2109.03201}, 2021.

\bibitem{kirillov2023segment}
A.~Kirillov, E.~Mintun, N.~Ravi, H.~Mao, C.~Rolland, L.~Gustafson, and R.~Girshick, ``Segment anything,'' in \emph{Proceedings of the IEEE/CVF International Conference on Computer Vision}, 2023, pp. 4015--4026.

\bibitem{wang2023sammed3d}
H.~Wang, S.~Guo, J.~Ye, Z.~Deng, J.~Cheng, T.~Li, and Y.~Shen, ``Sam-med3d: towards general-purpose segmentation models for volumetric medical images,'' \emph{arXiv preprint}, 2023.

\bibitem{ma2024segment}
J.~Ma, Y.~He, F.~Li, L.~Han, C.~You, and B.~Wang, ``Segment anything in medical images,'' \emph{Nature Communications}, vol.~15, no.~1, p. 654, 2024.

\bibitem{adams2023can}
J.~Adams and S.~Y. Elhabian, ``Can point cloud networks learn statistical shape models of anatomies?'' in \emph{International Conference on Medical Image Computing and Computer-Assisted Intervention}.\hskip 1em plus 0.5em minus 0.4em\relax Springer Nature Switzerland, 2023, pp. 486--496.

\bibitem{li2023medshapenet}
J.~Li, Z.~Zhou, J.~Yang, A.~Pepe, C.~Gsaxner, G.~Luijten, and M.~Reyes, ``Medshapenet--a large-scale dataset of 3d medical shapes for computer vision,'' \emph{arXiv preprint arXiv:2308.16139}, 2023.

\bibitem{qi2017pointnet}
C.~R. Qi, H.~Su, K.~Mo, and L.~J. Guibas, ``Pointnet: Deep learning on point sets for 3d classification and segmentation,'' in \emph{Proceedings of the IEEE conference on computer vision and pattern recognition}, 2017, pp. 652--660.

\bibitem{qi2017pointnet++}
C.~R. Qi, L.~Yi, H.~Su, and L.~J. Guibas, ``Pointnet++: Deep hierarchical feature learning on point sets in a metric space,'' in \emph{Advances in neural information processing systems}, vol.~30, 2017.

\bibitem{lin2024infocd}
F.~Lin, Y.~Yue, Z.~Zhang, S.~Hou, K.~Yamada, V.~Kolachalama, and V.~Saligrama, ``Infocd: a contrastive chamfer distance loss for point cloud completion,'' \emph{Advances in Neural Information Processing Systems}, vol.~36, 2024.

\bibitem{lin2024loss}
F.~Lin, H.~Liu, H.~Zhou, S.~Hou, K.~D. Yamada, G.~S. Fischer, Y.~Li, H.~K. Zhang, and Z.~Zhang, ``Loss distillation via gradient matching for point cloud completion with weighted chamfer distance,'' in \emph{2024 IEEE/RSJ International Conference on Intelligent Robots and Systems (IROS)}.\hskip 1em plus 0.5em minus 0.4em\relax IEEE, 2024, pp. 511--518.

\bibitem{yuan2018pcn}
W.~Yuan, T.~Khot, D.~Held, C.~Mertz, and M.~Hebert, ``Pcn: Point completion network,'' in \emph{2018 international conference on 3D vision (3DV)}.\hskip 1em plus 0.5em minus 0.4em\relax IEEE, 2018, pp. 728--737.

\bibitem{xia2021asfmnet}
Y.~Xia, Y.~Xia, W.~Li, R.~Song, K.~Cao, and U.~Stilla, ``Asfm-net: Asymmetrical siamese feature matching network for point completion,'' \emph{In Proceedings of the 29th ACM international conference on multimedia}, pp. 1938--1947, 2021.

\bibitem{yu2018punet}
L.~Yu, X.~Li, C.-W. Fu, D.~Cohen-Or, and P.-A. Heng, ``Pu-net: Point cloud upsampling network,'' \emph{In Proceedings of the IEEE conference on computer vision and pattern recognition}, pp. 2790--2799, 2018.

\bibitem{qian2021pugcn}
G.~Qian, A.~Abualshour, G.~Li, A.~Thabet, and B.~Ghanem, ``Pu-gcn: Point cloud upsampling using graph convolutional networks,'' in \emph{Proceedings of the IEEE/CVF Conference on Computer Vision and Pattern Recognition}, 2021, pp. 11\,683--11\,692.

\bibitem{he2023gradpu}
Y.~He, D.~Tang, Y.~Zhang, X.~Xue, and Y.~Fu, ``Grad-pu: Arbitrary-scale point cloud upsampling via gradient descent with learned distance functions,'' in \emph{Proceedings of the IEEE/CVF Conference on Computer Vision and Pattern Recognition}, 2023, pp. 5354--5363.

\bibitem{du2022cascaded}
H.~Du, X.~Yan, J.~Wang, D.~Xie, and S.~Pu, ``Point cloud upsampling via cascaded refinement network,'' in \emph{Proceedings of the Asian Conference on Computer Vision}, 2022, pp. 586--601.

\bibitem{zhao2021pointtransformer}
H.~Zhao, L.~Jiang, J.~Jia, P.~H. Torr, and V.~Koltun, ``Point transformer,'' \emph{In Proceedings of the IEEE/CVF international conference on computer vision}, pp. 16\,259--16\,268, 2021.

\bibitem{chen2021shape}
X.~Chen, N.~Ravikumar, Y.~Xia, R.~Attar, A.~Diaz-Pinto, S.~K. Piechnik, S.~Neubauer, S.~E. Petersen, and A.~F. Frangi, ``Shape registration with learned deformations for 3d shape reconstruction from sparse and incomplete point clouds,'' \emph{Medical image analysis}, vol.~74, p. 102228, 2021.

\bibitem{beetz2022point2mesh}
M.~Beetz, A.~Banerjee, and V.~Grau, ``Point2mesh-net: Combining point cloud and mesh-based deep learning for cardiac shape reconstruction,'' in \emph{International Workshop on Statistical Atlases and Computational Models of the Heart}.\hskip 1em plus 0.5em minus 0.4em\relax Springer, 2022, pp. 280--290.

\bibitem{beetz2023multi}
------, ``Multi-objective point cloud autoencoders for explainable myocardial infarction prediction,'' in \emph{International Conference on Medical Image Computing and Computer-Assisted Intervention}.\hskip 1em plus 0.5em minus 0.4em\relax Springer, 2023, pp. 532--542.

\bibitem{hu2021point}
B.~Hu, B.~Lei, Y.~Shen, Y.~Liu, and S.~Wang, ``A point cloud generative model via tree-structured graph convolutions for 3d brain shape reconstruction,'' in \emph{Pattern Recognition and Computer Vision: 4th Chinese Conference, PRCV 2021, Beijing, China, October 29--November 1, 2021, Proceedings, Part II 4}.\hskip 1em plus 0.5em minus 0.4em\relax Springer, 2021, pp. 263--274.

\bibitem{jiang2023cpaconv}
A.~Jiang, F.~Liu, Y.~Zhang, R.~Kong, F.~Luo, W.~C. Huang, and J.~N. Zou, ``Cpaconv-poco: a continuous position adaptive convolution based poco for lung nodule 3d reconstruction,'' in \emph{2023 IEEE International Conference on Bioinformatics and Biomedicine (BIBM)}.\hskip 1em plus 0.5em minus 0.4em\relax IEEE, 2023, pp. 1166--1173.

\bibitem{chen2024cartilage}
J.~Chen, F.~Guo, and Q.~Tang, ``Cartilage repair based on point cloud completion,'' in \emph{2024 5th International Conference on Computer Vision, Image and Deep Learning (CVIDL)}.\hskip 1em plus 0.5em minus 0.4em\relax IEEE, 2024, pp. 955--959.

\bibitem{xzhang2023anatomical}
X.~Zhang, Y.~Liu, S.~Ali, X.~Zhao, M.~Sun, M.~Han, T.~Liu, P.~Zhai, Z.~Cui, P.~Zhang \emph{et~al.}, ``Anatomical-aware point-voxel network for couinaud segmentation in liver ct,'' in \emph{International Conference on Medical Image Computing and Computer-Assisted Intervention}.\hskip 1em plus 0.5em minus 0.4em\relax Springer, 2023, pp. 465--474.

\bibitem{xzhang2024robust}
X.~Zhang, S.~Ali, T.~Liu, X.~Zhao, Z.~Cui, M.~Han, S.~Ma, J.~Zhu, Y.~Kang, L.~Wang \emph{et~al.}, ``Robust and smooth couinaud segmentation via anatomical structure-guided point-voxel network,'' \emph{Computers in Biology and Medicine}, vol. 182, p. 109202, 2024.

\bibitem{yassin2024medshapenet}
A.~Yassin, G.~Luijten, A.~Elsakka, A.~Ferreira, B.~Puldai, V.~Alves, and J.~Egger, ``A medshapenet foundation model,'' 2024.

\bibitem{zhang2025hierarchical}
G.~Zhang, J.~Yang, and Y.~Li, ``Hierarchical feature learning for medical point clouds via state space model,'' \emph{arXiv preprint arXiv:2504.13015}, 2025.

\bibitem{cai2024statistical}
X.~Cai, Y.~Wu, J.~Huang, L.~Wang, Y.~Xu, and S.~Lu, ``Application of statistical shape models in orthopedics: a narrative review,'' \emph{Intelligent Medicine}, vol.~4, no.~4, pp. 249--255, 2024.

\bibitem{guo2024tetsphere}
M.~Guo, B.~Wang, K.~He, and W.~Matusik, ``Tetsphere splatting: Representing high-quality geometry with lagrangian volumetric meshes,'' \emph{arXiv preprint arXiv:2405.20283}, 2024.

\bibitem{botsch2010polygon}
M.~Botsch, L.~Kobbelt, M.~Pauly, P.~Alliez, and B.~L{\'e}vy, \emph{Polygon mesh processing}.\hskip 1em plus 0.5em minus 0.4em\relax CRC press, 2010.

\bibitem{chen2022ndc}
Z.~Chen, A.~Tagliasacchi, T.~Funkhouser, and H.~Zhang, ``Neural dual contouring,'' \emph{ACM Transactions on Graphics (TOG)}, vol.~41, no.~4, pp. 1--13, 2022.

\bibitem{liu2021deep}
P.~Liu, H.~Han, Y.~Du, H.~Zhu, Y.~Li, F.~Gu, H.~Xiao, J.~Li, C.~Zhao, L.~Xiao \emph{et~al.}, ``Deep learning to segment pelvic bones: large-scale ct datasets and baseline models,'' \emph{International Journal of Computer Assisted Radiology and Surgery}, vol.~16, pp. 749--756, 2021.

\bibitem{yifan2019patch}
W.~Yifan, S.~Wu, H.~Huang, D.~Cohen-Or, and O.~Sorkine-Hornung, ``Patch-based progressive 3d point set upsampling,'' in \emph{Proceedings of the IEEE/CVF Conference on Computer Vision and Pattern Recognition}, 2019, pp. 5958--5967.

\bibitem{zhang2024rethinking}
T.~Zhang, ``Rethinking data input for point cloud upsampling,'' in \emph{Proceedings of 17th International Conference on Machine Learning and Computing}, L.~Huang and D.~Greenhalgh, Eds.\hskip 1em plus 0.5em minus 0.4em\relax Cham: Springer Nature Switzerland, 2025, pp. 29--41.

\bibitem{ZHANG2025104467}
\BIBentryALTinterwordspacing
T.~Zhang and B.~Wang, ``Representation learning of point cloud upsampling in global and local inputs,'' \emph{Computer Vision and Image Understanding}, vol. 260, p. 104467, 2025. [Online]. Available: \url{https://www.sciencedirect.com/science/article/pii/S1077314225001900}
\BIBentrySTDinterwordspacing

\bibitem{lorensen1998marching}
W.~E. Lorensen and H.~E. Cline, ``Marching cubes: A high resolution 3d surface construction algorithm,'' in \emph{Seminal graphics: pioneering efforts that shaped the field}, 1998, pp. 347--353.

\bibitem{cheng2020morphing}
Q.~Cheng, P.~Sun, C.~Yang, Y.~Yang, and P.~X. Liu, ``A morphing-based 3d point cloud reconstruction framework for medical image processing,'' \emph{Computer methods and programs in biomedicine}, vol. 193, p. 105495, 2020.

\bibitem{xi2021recovering}
L.~Xi, Y.~Zhao, L.~Chen, Q.~H. Gao, W.~Tang, T.~R. Wan, and T.~Xue, ``Recovering dense 3d point clouds from single endoscopic image,'' \emph{Computer Methods and Programs in Biomedicine}, vol. 205, p. 106077, 2021.

\bibitem{yang20243d}
Z.~Yang, J.~Dai, and J.~Pan, ``3d reconstruction from endoscopy images: A survey,'' \emph{Computers in biology and medicine}, vol. 175, p. 108546, 2024.

\end{thebibliography}


\end{document}